\documentclass[]{ourstyle}
\usepackage[table]{xcolor}
\usepackage{amssymb}
\usepackage{dsfont}
\usepackage{algorithm}
\usepackage{algpseudocode}

\usepackage{caption}
\usepackage{svg}
\usepackage{graphicx}
\usepackage{xparse}

\usepackage[table,dvipsnames]{xcolor}
\usepackage{graphicx}
\usepackage{multirow}
\usepackage{pifont} 

\usepackage{tcolorbox}

\usepackage{enumitem}

\definecolor{verylightblue}{HTML}{f4fbff}
\definecolor{lightblue}{HTML}{77C9FF}
\definecolor{mediumblue}{HTML}{0099FF}

\definecolor{verylightred}{HTML}{fffbf4}
\definecolor{lightred}{HTML}{FF5957}
\definecolor{mediumred}{HTML}{FF5900}

\tcbset{
  aibox/.style={
    width=480pt,
    top=7pt,
    bottom=5pt,
    colback=metabg,
    colframe=black,
    colbacktitle=black,
    enhanced,
    center,
    attach boxed title to top left={yshift=-0.1in,xshift=0.15in},
    boxed title style={boxrule=0pt,colframe=white,},
  }
}
\newtcolorbox{AIbox}[2][]{aibox,title=#2,#1}

\usepackage{xcolor} 
\usepackage{tikz}
\usepackage{ulem}
\usepackage{cancel}
\usepackage{amsmath}
\usepackage{hyperref} 
\usetikzlibrary{fit,calc}

\DeclareRobustCommand{\xsectionj}[1]{%
  $\xcancel{\text{#1}}$%
}

\DeclareRobustCommand{\xsectionk}[1]{%
  \vspace{2mm}$\xcancel{\text{#1}}$\vspace{1mm}%
}

\colorlet{mypink}{red!30}
\colorlet{myblue}{orange!30}
\colorlet{mypurple}{green!10}

\definecolor{customblue}{HTML}{0099ff}

\definecolor{customred}{HTML}{ff006b}

\title{\xsectionk{Self-Improving AI}\vspace{-1.8cm}~~\\
\hspace{0.8cm} ~~~~~~~~~~~~~~~~~~~~~~~~~~~~~~~~~~~~~~~~AI \& Human Co-Improvement\\
\vspace{0.1cm}\hspace{6cm}~~~for Safer Co-Superintelligence}

\affiliation{FAIR at Meta}
\author{Jason Weston}
\author{Jakob Foerster}

\abstract{
Self-improvement is a goal currently exciting the field of AI, but is fraught with danger, and may take time to fully achieve. We advocate that a more achievable and better goal for humanity is to maximize co-improvement: collaboration between human researchers and AIs to achieve co-superintelligence. 
That is, specifically targeting improving AI systems' ability to work with human researchers to conduct AI research together, from ideation to experimentation, in order to  both accelerate AI research and to generally endow both AIs and humans with 
safer superintelligence
through their symbiosis.
Focusing on including human research improvement in the loop will both get us there faster, and more safely.  
}

\begin{document}

\maketitle

\section*{\xsectionj{The quest for Self-Improving AI}}

\label{section:intro}


AI that improves itself has been the primary goal of the field 
since its inception \citep{turing2021computing}. Historically, 
practical instantiations 
have focused on parameterizing a model in terms of weights, and then finding the best choice of those weights, without human intervention -- from linear models through to neural networks. The 2010s onwards marked the era of scaling to ever-larger models \citep{sutskever2014sequence}, yielding large performance gains, but still focused on self-improving via weights only, while the architecture, data, objective function, update rule, and implementation (code) were mostly fixed.
The current era has expanded that self-improvement search to improving all aspects through learning:  models that create their own training data  \citep{wang2022self}, challenge themselves to be better \citep{zhou2025self,zhao2025absolute}, and  learn to evaluate and reward themselves based on their own performance on these tasks \citep{bai2022constitutional,yuan2401self,wang2024self}; see \autoref{tab:self_improvement_types}.  
Some of these axes have already provided sizable gains in performance, and approaches like synthetic data creation and 
LLM-as-a-Judge 
are now standard building blocks for frontier models. 
The quest for AI that improves its own architectures and rewrites its own code is still in its infancy \citep{schmidhuber2007godel,ren2021comprehensive}, but early signs show its promise, with a current push on autonomous AI research agents \citep{lu2024ai,starace2025paperbench,chan2024mle,audran2025does,nathani2025mlgym}.

It seems clear by now that we are marching towards ever more intelligent AI systems that in the long run will surpass humans in all task metrics, and by a large margin. Fully realized self-improvement is clearly an end-game marker.
However, endowing AIs with this autonomous ability  without 
appropriate 
guidance built into the system is fraught with danger for humankind -- from misuse through to  misalignment \citep{hendrycks2023overview,openai2025preparedness}. 
Nevertheless,  there is still time left before  AI eclipses humans in all endeavors, and in AI research in particular. 
We thus propose that it is better to focus on humans and AI working together to solve these problems.

\section*{A better quest for humans: Co-Improving AI}

Our central position is that 
{{\textbf{``Solving AI'' is accelerated by building  AI that collaborates with humans to solve AI}}}.
This is distinct from the goal of self-improving AI, which seeks to eliminate humans from the loop as quickly as possible, where AI performs its own research and learning autonomously.
Instead, we advocate for \textbf{co-improvement}, whereby 
 collaborative AI agents are built with the goal of conducting research {\it{with}} humans. Thus, we accelerate the research 
 \textit{with the research}.
Importantly, including humans in the loop 
allows us the ability 
to steer the research 
in the right directions, 
i.e. ``Solving AI'' means  a \textit{positive solution for humanity}. 
In particular, we believe that such a \textit{positive solution} is one where AI augments and enables humans in all areas of society, rather than pursuing full automation that removes human decision-making.

 While today most of the AI research is done by humans, we expect over time the burden will become more shared -- as AI improves it can work with us to take more of the workload in providing solutions.
 Collaboration can take advantage of the complementary skill sets of humans and AI, which currently excel in quite different areas, while we expect over time AI will continue to outpace us in more and more dimensions. 
However, because AI is not mature enough  yet  to fully self-improve and is  susceptible to misalignment, we expect that co-improvement will get us there faster \textit{and} more safely. 
That is, with the help of AI we are more likely to solve the capability and safety problems of AI --- but with humans in the loop, collaborating on the research. Thus, co-improvement can help induce positive outcomes for humanity.

\begin{figure}[t!]
\centering
\begin{tcolorbox}[colback=Lavender!20,colframe=Lavender!10, boxrule=0.8pt, arc=6pt]
\centering
\includegraphics[width=13cm,trim=2.5cm 4cm 3cm 4cm, clip]{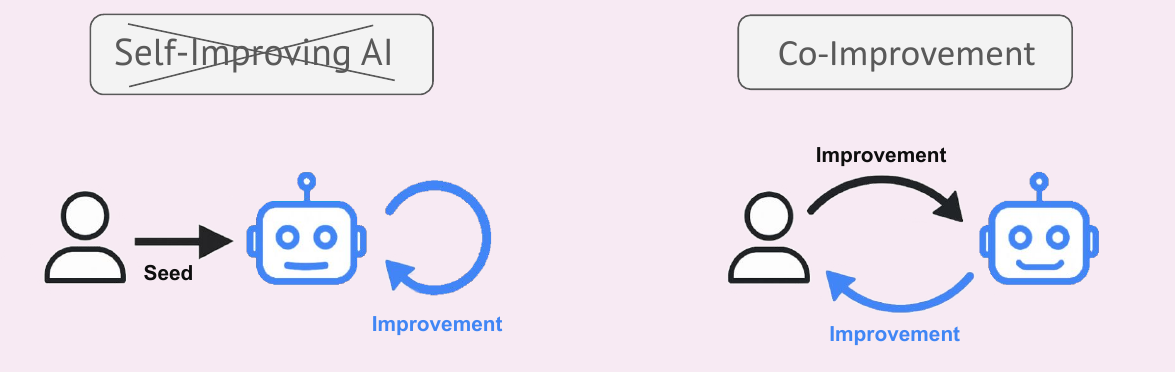}
\caption*{\hspace{-1.8mm}In {self-improving AI}, humans build an AI that improves itself autonomously with no human in the loop.
In \textbf{co-improvement} humans build the first system, but then collaborate with the AI for further improvement, so both humans and AI get the benefit. Co-improvement is a bidirectional collaboration between humans and AI where each improves the other’s ability \& understanding over time. 
Unlike self-improvement, which focuses on autonomous model updates, co-improvement centers on joint progress: humans help the AI achieve greater abilities, while the AI augments human cognition, research progress, and general capabilities outside of research.}
\end{tcolorbox}
\end{figure}
\paragraph{\textbf{Aren't we already doing this?}}

There are multiple plausible pathways to superintelligence, and current AI systems are already helping us explore these research directions to some degree, e.g. through code assistance and writing assistance \citep{chen2021evaluating,ma2023ai,lee2022coauthor,padmakumar2023does}. 
And, in general, improving the overall capabilities of frontier models does equip them with some of the skills suitable for research collaboration -- as a byproduct \citep{radford2019language,brown2020language}. However, we typically see that skills improve further when they are targeted, for example lots of effort has been spent on improving AI coding skills, resulting in improved AI coding ability. However, while coding isn't solved either \citep{tie2024llms}, there is much more to ``solving'' AI than coding alone.  Our central point is that with more development resources spent on endowing AI with AI research collaboration skills, these abilities will get better too.

\newcommand{\rotcell}[1]{\rotatebox[origin=b]{90}{\footnotesize #1}}

\newcommand{\colA}{\cellcolor{CornflowerBlue!10}} 
\newcommand{\colB}{\cellcolor{Apricot!18}}        
\newcommand{\colC}{\cellcolor{Green!12}}          
\newcommand{\colD}{\cellcolor{Lavender!20}}       

\newcommand{\good}{\(\bullet\)}
\newcommand{\bad}{\(\circ\)}

\begin{table*}[h!]
\centering

\caption{\textbf{Co-improvement} goals across major AI development activities, in order to achieve 
co-superintelligence.
\label{tab:co_improvement_types}
}
\vspace{-2mm}
{
\begin{minipage}[t]{0.48\linewidth}
\centering

\renewcommand{\arraystretch}{1.28}

\definecolor{ApricotSoft}{HTML}{F8D9BF}     
\definecolor{CoralTint}{HTML}{FFD6C8}       
\definecolor{Cream}{HTML}{FFF4E8}           
\definecolor{BlushGray}{HTML}{F1E8E3}       
\definecolor{HeaderCream}{HTML}{FCF6EE} 

\begin{footnotesize}
\begin{tabular}{
    >{\colA\arraybackslash\raggedright}p{2.5cm}|
    >{\colD\arraybackslash}p{4.8cm}|
}
\rowcolor{HeaderCream}
\textbf{Category} &
\textbf{Mechanism} \\
\hline

\textbf{Collaborative problem identification} &
Humans and AI help jointly define goals; identify current failures, brainstorm, propose unexplored directions, take into account existing work. \\
\hline
\textbf{Benchmark creation \& problem evaluation} &
Given identified problem, jointly define desiderata; benchmark construction \& performance analysis;  refine benchmarks to validate problem. \\
\hline

\textbf{Method innovation \& idea generation} &
Jointly brainstorm and identify solutions: systems, architectures, algorithms, training data, recipes and overall code designs for future models.\\
\hline

\textbf{Joint experiment design} &
Co-design overall plan to test innovations: experiment protocol and setting, further benchmark identification, proposed ablations, \dots \\
\hline

\textbf{Collaborative execution} &
Humans and AI co-produce  and run multi-step workflows (implementation, experiments).\\
\hline

\textbf{Evaluation \& error analysis} &
Analyzing performance both on benchmarks and individual cases for successes \& failures (at scale via AI);  feedback loop for research iteration.\\

\hline

\end{tabular}
\end{footnotesize}

\end{minipage}
\hfill
\begin{minipage}[t]{0.48\linewidth}
\centering

\renewcommand{\arraystretch}{1.28}

\definecolor{ApricotSoft}{HTML}{F8D9BF}     
\definecolor{CoralTint}{HTML}{FFD6C8}       
\definecolor{Cream}{HTML}{FFF4E8}           
\definecolor{BlushGray}{HTML}{F1E8E3}       
\definecolor{HeaderCream}{HTML}{FCF6EE}

\begin{footnotesize}
\begin{tabular}{
    >{\colA\arraybackslash\raggedright}p{2.5cm}|
    >{\colD\arraybackslash}p{4.8cm}|
}
\rowcolor{HeaderCream}
\textbf{Category} &
\textbf{Mechanism} \\
\hline

\textbf{Safety \& alignment} &
Humans and AI co-develop methods as well as values, 
and constitutions. 
Use whole research cycle listed (left) to develop and test them.\\

\hline

\textbf{Systems \& infrastructure co-design} &
Jointly architecting pipelines, optimizations, configs, and reproducibility improvements. \\
\hline

\textbf{Integrating into real-world systems} &
Collaborating to convert research to real-world use; in turn this may identify  further required research. \\
\hline

\textbf{Scientific communication } &
Jointly drafting documentation \& writeups, including  figures and results; ensuring clarity of message and correctness. \\
\hline

\textbf{Collective intelligence \& group research} &
Multi-human \& AI collaboration for given problem; aggregate viewpoints, structure debate, and synthesize consensus \& actionable steps. \\
\hline

\textbf{Bidirectional co-improvement } &
Overall collaboration aims to enable increased intelligence in both humans \& AI, including all manifested learnings from the research cycle, with the goal of achieving co-superintelligence.
\\
\hline

\end{tabular}
\end{footnotesize}

\end{minipage}
}
\centering
\vspace{8mm}
\caption{Major interpretations of {\textbf{self-improvement}} in AI research 
together with outstanding research challenges.}
\label{tab:self_improvement_types}
\renewcommand{\arraystretch}{1.28}

\begin{footnotesize}
\vspace{-2mm}
\begin{tabular}{
    >{\colA\raggedright\arraybackslash}p{1.9cm}|                    
    >{\columncolor{Gray!10}\centering\arraybackslash}p{0.035cm}      
    >{\columncolor{Gray!18}\centering\arraybackslash}p{0.035cm}      
    >{\columncolor{Gray!10}\centering\arraybackslash}p{0.035cm}      
    >{\columncolor{Gray!18}\centering\arraybackslash}p{0.035cm}      
    >{\columncolor{Gray!10}\centering\arraybackslash}p{0.035cm}|     
    >{\colB}p{3.8cm}|                                               
    >{\colC}p{3.0cm}|                                               
    >{\colD}p{3.8cm}                                                
}
\hline
& \multicolumn{5}{>{\columncolor{Gray!14}}c|}{\textbf{Learnable axis}} 
& & & \\
\textbf{Category} &
\rotcell{{Parameters}} &
\rotcell{{Train data}} &
\rotcell{{Objective}} &
\rotcell{{Architecture}} &
\rotcell{{Overall code}} &
\textbf{Mechanism} &
\textbf{Representative Examples} &
\textbf{Open Issues / Research Directions} \\
\hline
\textbf{Classic parameter optimization} &
\good & \bad & \bad & \bad & \bad &
Model updates its own weights given data \& fixed objective, i.e., classic training algorithms. &
Gradient descent, SFT, LLM pre-training. &
Data inefficiency; out-of-distribution generalization; compute inefficiency (model size, data size). \\
\hline

\textbf{Self-training on own generations or actions} &
\good & \good & \bad & \bad & \bad &
Model learns from its own generated outputs and 
reasoning traces; could use tools/actions in an environment. &
RL with fixed/verifiable rewards \citep{sutton1998reinforcement,mnih2013playing,silver2017mastering, silver2018general, vinyals2019grandmaster, berner2019dota, luong2024reft, pang2024iterative,
guo2025deepseek}; STaR \citep{zelikman2022star}. &
Computational efficiency; reward hacking; can drift away from human reasoning without strong priors. \\
\hline

\textbf{Self-challenging / Self-play \& Synthetic data creation} &
\good & \good & \bad & \bad & \bad &
Model improves via self-generated tasks; can simultaneously improve  its ability to generate tasks. &
(CoT-)-Self-Instruct \citep{wang2022self,yu2025cot}, 
Self-Challenging Agents \citep{zhou2025self}, Absolute Zero \citep{zhao2025absolute}, SPICE \citep{liu2025spice}.  &
Task quality \& correctness, 
diversity \& generalization beyond synthetic tasks.\\
\hline

\textbf{Self-evaluation / Self-reward} &
\good & \good & \good & \bad & \bad &
Model generates and applies its own feedback or reward signals; can learn reward model. &
Self-Rewarding \citep{yuan2401self}, Self-refinement, RLAIF \citep{lee2023rlaif}, Constitutional AI \citep{bai2022constitutional}. &
Ensuring value alignment; validating reliability of self-generated judgments. \\
\hline


\textbf{Algorithm or architecture self-modification} &
\good & \bad & \good & \good & \good &
Model changes its own structure, code, compiler transforms, or learning algorithm. &
Neural arch. search \citep{elsken2019neural}; code optimization, AlphaEvolve \citep{novikov2025alphaevolve}; ``AI scientist'' \citep{lu2024ai,starace2025paperbench,nathani2025mlgym}. &
Efficiency; ensuring safety and correctness; interpretability of modifications. \\ 
\hline

\textbf{Recursive self-improvement} &
\good & \good & \good & \good & \good &
System designs successively more capable versions of itself, potentially achieving AGI/SI. & 
Gödel Machine \citep{schmidhuber2007godel}; theoretical AGI discussions \citep{yudkowsky2008artificial}. &
Executing on this currently theoretical vision; ensuring safety.\\
\hline

\textbf{Metaphorical or rhetorical usage} &
? & ? & ?  & ? & ? &
Non-technical use implying progressive capability or autonomy. &
Media or industry narratives. &
Clarifying terminology; avoiding 
conceptual confusion. \\
\hline

\end{tabular}
\end{footnotesize}
\end{table*}

\paragraph{\textbf{What can we gain?}}

Progress in AI has been made with a combination of both training data and method changes from architecture through to training objectives, often with these advances working in tandem, leading to notable paradigm shifts. For example, the creation of Imagenet and the introduction of AlexNet \citep{deng2009imagenet,krizhevsky2012imagenet}, curating web data and scaling transformers \citep{vaswani2017attention,devlin2019bert,radford2018improving},
the labeling of instruction following data and building of RLHF training  \citep{christiano2017deep,schulman2017proximal,ouyang2022training}, or the collection of verifiable reasoning tasks and 
the use of RLVR for training chain-of-thought \citep{cobbe2021training,hendrycks2021measuring,zhu2024deepseek,guo2025deepseek}.
In each case it took human researchers significant effort, with many smaller intermediate results as well as wrong directions and dead ends,  in order to find these wins. Any improvement in our ability to do research will speed up this process. 
Hence, co-research with strong AI systems built to collaborate with us should accelerate finding the unknown new paradigm shifts which are currently missing.

Overall, we expect co-improvement can provide: (i)  faster progress to find important paradigm shifts; (ii) more transparency and steerability than direct self-improvement in making this progress; (iii) more focus on human-centered safe AI. For example,  we may be able to develop systems that are super-human at ML theory, so we could have provably safe AI. In contrast, an entirely autonomous AI self-improvement system can suffer from goal misspecification (e.g. what it means to "solve AI" does not take human needs into account).

\paragraph{\textbf{How do we do it?}} 

In order to build AI that can collaborate with us on research, we should put some of our focus on building AI possessing these skills.
So, that means measuring the research collaboration skills of AI with new benchmarks, and constructing training data and methods that improve these benchmarks, much as we do with building other skills. These skills should cover all major AI research activities that comprise the end-to-end research pipeline. We define some major ones in
\autoref{tab:co_improvement_types}.
These include collaboration with us to identify research problems, 
create training data and benchmarks, innovate methods, design and execute experiments, and conduct evaluation and error analysis which is then fed back to refine the whole process.
Similarly, co-design and co-development of safety \& alignment,  systems improvements, integration of these innovations into real-world use cases, and scientific communication are also important goals. Crucially, in contrast to recent end-to-end AI scientist approaches~\cite{lu2024ai,nathani2025mlgym} the goal is to improve research quality, rather than to accelerate research artifact production (such as papers) via full automation.

\paragraph{\textbf{From co-improvement to co-superintelligence}} 

We envision that the first goal of co-improvement is to improve our ability to conduct research into improving AI. We expect the end result when successful, 
like in the self-improving paradigm, will be a superintelligence system with ability to self-improve. However, the difference is when humans are working with the AI system to help achieve this at each step of the loop, we  have more opportunities to guide this process towards positive benefits to humanity.
In particular, we can consider safety and societal harms  (see later section), and increase the collective knowledge of humanity at each step.

Going forward, we further envision that the goals of co-improvement could
shift from building AI that collaborates on AI research toward co-improvement on \textit{all kinds} of research or important topics to humanity. As AI becomes ever more capable, these new skills hopefully become even easier to attain. From the human societal standpoint, building AI can help \textit{humans improve themselves, their abilities and knowledge, and their situation}. We can thus focus on building AI towards those goals. We thus refer to AI helping us achieve these abilities, beyond our current ones, as \textbf{co-superintelligence}, emphasizing what AI can give back to humanity.

\paragraph{\textbf{Co-improvement and societal harms \& benefits}} 

As capabilities increase the potential for harms can increase. Today, there are many harms due to models {\textit{not being capable enough}}, for example jailbreaking \citep{wei2023jailbroken} occurs because the models do not ``understand'' they are jailbroken. Collaborating with AI can help find research solutions to fix these problems -- their own problems! -- i.e., find and implement new capabilities that lead to safer models, new safety procedures,  and co-developing values, constraints, and constitutions. This optimistic viewpoint contends that AIs increased capabilities can thus be leveraged to decrease harms -- if done correctly.

As AI become more capable there is also  an optimistic opportunity to help with many other societal issues beyond their own impact. Rather than the self-improved superintelligence dystopian paradigm where an AI overlord  dictates best practices to humans, the co-improvement paradigm suggests 
collaborative help to synthesize consensus and find actionable steps to solve problems. Multi-human and AI collaboration could help aggregate viewpoints, structure debate and help humans come to positive conclusions and outcomes. 

\paragraph{\textbf{Co-improvement and openness}}

If humanity wants to improve its scientific knowledge, the clearest way to do that is to use the scientific method. This  means conducting 
reproducible science with openly disseminated results, so others can verify or build on them, and collective knowledge can advance.
Co-improvement can help this knowledge advance more quickly in both the fields of AI, and potentially any other field of science as well.
We note there is a current shift away from open AI research from a number of industrial labs. As stated in \cite{brundage2018malicious} we agree that {\textit{``concerns about misuse should not be used as an excuse to reduce openness to a greater extent than is required, for instance, when the real motivation is about corporate competitiveness.''}}
Nevertheless in AI, as in other fields of science,  we believe {\textit{managed openness}} should be considered in order to combat societal harms when required  
\citep{brundage2018malicious}, which should be an ongoing discussion as capabilities increase.

\section*{Relation to other existing positions}

\paragraph{\textbf{Related Positions}} 
Our position is related to treatise on human-centered AI \citep{shneiderman2022human,wilder2020learning,horvitz2007reflections,Zuckerberg2025}, but in our case more specifically for the goal of achieving (co-)superintelligence via collaborative research.
Similarly, the work of \citep{dafoe2021cooperative} advocates for cooperative AI, and finding common ground with machines, where research would be a special case. The recent AI co-scientist  \citep{gottweis2025towards} is a related practical system, but for the biomedical domain, not for AI research.
Nevertheless, we agree that work on all forms of human-AI collaboration is also an important goal for future human society with the advent of superhuman AIs. Other works 
emphasize that misalignment is a challenging goal, and
advocate for the importance of AI's objectives to be human-oriented \citep{russell2022human}. 
Scalable oversight is a related solution of allowing humans to intervene in the alignment process by providing supervision \citep{bowman2022measuring}.  
Importantly, our position is that this challenging problem 
can and should be tackled by research collaboration, which is bidirectional. 
For example, collaboration with AI may more readily help find flaws in their own designs, which also shares common goals with \citep{bengio2025safer}. Overall, we believe this research should be in progress now -- not added on later when systems are fully deployed and it is too late.

\paragraph{\textbf{Contrasting Positions}}

Various works have described autonomous self-improvement and possible ways it can be achieved, for example \citep{schmidhuber2007godel,clune2019ai,silver2021reward,silver2025era,bostrom2024superintelligence}. 
Frontier labs are clearly aiming to develop self-improving AI, whilst also knowing that it has serious long-term risks \citep{openai2025preparedness,bechard2025howclose,aschenbrenner2024situational}.
Correspondingly, a large number of works have also developed practical instantiations with various technical contributions, see \autoref{tab:self_improvement_types} for examples. 

The authors of \citep{silver2025era} advocate for ``an era of experience'' where self-improvement is obtained by autonomous learning from the AI's own experience. The implication is that there is little cooperation with humans, for example they write that AIs will \textit{``autonomously design and conduct experiments in fields like materials science, medicine, or hardware design''}. They also admit that this \textit{``provides fewer opportunities for humans to intervene and mediate the agent’s actions, and therefore requires a high bar of trust and responsibility''}.
Others see little role for humans when the self-improvement goal is achieved, e.g. from \citep{Schmidhuber2016}: \textit{``\dots AI will colonise the galaxy. 
Humans are not going to play a big role there, but that’s ok. We should be proud of being part of a grand process that transcends humankind.''}

In contrast, we envision a world where humans are always a necessary, but maximally augmented, part of not only economic, scientific, but also all types of decision-making processes. We believe the AI community should fully embrace and implement this vision in our own pursuit of this long term goal.

\section*{Conclusion}

We have argued that the existing goal of autonomous self-improving AI is misguided for two reasons:  it is not the fastest way to achieve superintelligence, nor the safest. We advocate instead for co-improvement, whereby we (human researchers) focus on building AI that is collaborative 
and in particular collaborates with us to conduct research
-- to help us achieve ever more collaborative, capable and safer AI, with its help. 
Achieving this goal in turn could unlock the creation of extremely capable AIs in the future that can collaborate with us to solve the important goals and societal problems for humanity at large.

\section*{Acknowledgements}

This position paper, like the co-improvement paradigm we advocate for, was made with the help of AI. We also thank the humans at the AIRA and RAM teams in FAIR for useful discussions.

\bibliographystyle{unsrt}
\bibliography{paper}

\end{document}